\documentclass[review]{elsarticle}

\journal{Physica D}

\usepackage{amssymb}
\usepackage{amsmath}
\newcommand{\R}{\mathbb{R}}
\newcommand{\Z}{\mathbb{Z}}
\newcommand{\eps}{\varepsilon}
\newcommand{\tr}{\mbox{tr}}
\newcommand{\diag}{\mbox{diag}}

\begin{document}

\begin{frontmatter}

\title{Inference of modes for linear stochastic processes}

\author{R.S.MacKay} 
\address{Centre for Complexity Science and Mathematics Institute, University of Warwick, Coventry CV4 7AL, UK}
\ead{R.S.MacKay@warwick.ac.uk}




\begin{abstract}
For dynamical systems that can be modelled as asymptotically stable linear systems forced by Gaussian noise, this paper develops methods to infer or estimate their modes from observations in real time.  The modes can be real or complex.  For a real mode, we wish to infer its damping rate and mode shape.  For a complex mode, we wish to infer its frequency, damping rate and (complex) mode shape.  Their amplitudes and correlations are encoded in a mode covariance matrix.  The work is motivated and illustrated by the problem of detection of oscillations in power flow in AC electrical networks. Suggestions of other applications are given.
\end{abstract}


\end{frontmatter}


In memory of Professor Sir David John Cameron MacKay FRS (22 April 1967 -- 14 April 2016).

\section{Introduction}

\begin{figure}[htbp] 
   \centering
   \includegraphics[width=4.5in]{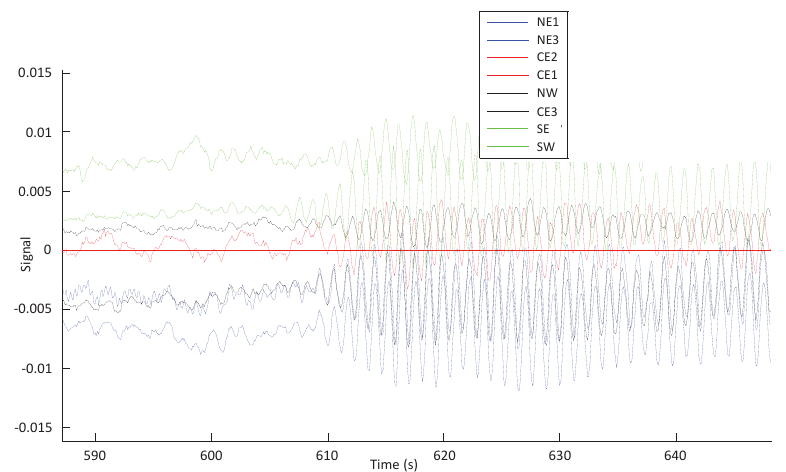} 
   \caption{Voltage angle at 7 locations in England relative to an 8th (CE2) as a function of time.  Reproduced with permission from \cite{TR+}.
   Angle differences drive power flow, so oscillations in angle differences indicate oscillations in power flow.}
   \label{fig:TR+}
\end{figure}

In January 2015, National Grid asked if I could improve their methods for detection of oscillations in power flow, to estimate frequency, damping constant, mode shape and amplitude.  Figure~\ref{fig:TR+} shows an example where such a mode of oscillation became clear, but National Grid want to detect them before they get excited enough to become clear, so that they can design and install suitable controllers to limit them.  This type of oscillation is called ``inter-area"; for a review of oscillations in electrical power flow, see \cite{P}.

As my brother David, author of \cite{M2}, was expert in data analysis, I asked him what he would recommend.  He responded ``Use a Gaussian process".  It looked a good idea and this paper is the result, though I have specialised to the class of linear stochastic processes, as I will explain.

A {\em Gaussian process (GP)} on a set $T$ is a probability distribution for functions $F: T \to \R$ such that for all $n\ge 1$ the marginal density $P$ for the vector of values $f_1,\ldots f_n = F(t_1),\ldots F(t_n)$ at any finite sequence $t_1,\ldots t_n \in T$ is Gaussian.  Examples for the set $T$ are $\R$ representing time, or the set $V$ of vertices in a graph representing spatial locations in a network, or $\R\times V$ for time and vertices, or $\R\times V \times I$ where $I$ is a set of labels representing components of a vector of values at each vertex and time.

A basic theorem (e.g.~\cite{RW}) for a GP is that there is a ``mean" function $M: T \to \R$ and a positive-definite ``covariance" function $C: T\times T \to \R$ such that
\begin{equation}
P(f_1,\ldots f_n) = (2\pi)^{-n/2} (\det c)^{-1/2} e^{-\frac12(f-m)^Tc^{-1}(f-m)},
\end{equation}
where $m$ is the vector with components $m_i = M(t_i)$ and $c$ is the matrix with components $c_{ij} = C(t_i,t_j)$.  $C$ being positive-definite means that for all $n\ge 1$, $t_1,\ldots t_n \in T$ and $v_1,\ldots v_n \in \R$ not all zero then $v^Tcv>0$.

It is convenient to extend the concept of GP to degenerate cases by allowing $C$ to be positive semi-definite (psd) ($v^T cv \ge 0$).  In this case $c$ may fail to be invertible but the above formula for the density $P$ can be understood as the product of a delta-function on the null space of $c$ and a Gaussian of complementary dimension on the range of $c$, centred at $m$.

Given a GP and observations of a realisation of it at a subset $T' \subset T$, possibly with an assumed Gaussian distribution for measurement error (essential if the covariance is not positive-definite), then Bayesian inference produces a posterior probability distribution for the realisation, and the calculation is just linear algebra.  

Given a family of GPs, labelled by one or more parameters, a prior probability distribution on the parameter space, and observations of a realisation, then Bayesian inference gives a posterior probability distribution over the joint space of parameters and realisations.  In particular, its marginal over the parameter space gives a posterior probability over the parameter space.  In general this can not be computed explicitly, but search algorithms can find the parameter values maximising the posterior likelihood.  In this way, one can infer the parameters.  Computational methods can also give an idea of the posterior uncertainty in the parameters.

There are many introductions to Gaussian processes, e.g.~\cite{M, RW, RR, RO+, L+}, and software packages to implement them and infer from them, e.g.~GPML.

Much GP modelling, however, seems to me to be ad hoc.  A family of covariance functions is chosen, for example to reflect assumed smoothness class or periodicity, the mean function is often set to zero, and a best fit to the data is obtained.  Instead, it seems to me better to use known or assumed structure of the system under study to choose a sensible class of models.  This strategy is recognised under the names ``hybrid modelling" or ``latent force models", e.g.~\cite{ALL}.

For time-dependent systems, in many contexts a natural class of models is an asymptotically stable continuous-time linear system forced by Gaussian noise.  Furthermore, it is often natural to assume the linear system to be autonomous (some say ``time-invariant") and the noise to be stationary, at least on short time-scales.  The noise is not necessarily white.  I make the assumption that it is the result of forcing some other autonomous asymptotically stable linear system with white Gaussian noise.  The noise model can be criticised but for electricity networks, load appears to be very close to Gaussian \cite{TR+}, and it is plausible that power imbalance is the result of first-order filtering Gaussian white noise, as will be discussed in Section~\ref{sec:ACelec}.  The end-result of the assumption on the noise is a skew-product asymptotically stable linear system (consisting of the real system and the noise filter) forced by Gaussian white noise.  

Another name for linear stochastic systems is continuous-time vector autoregressive (VAR) processes (see App.~B.2.1 of \cite{RW}).
Classic books on the discrete-time version of such models, including inference for them, are \cite{Ca,WH}.  In the latter, they are called Dynamic Linear Models.

Examples of this class of model are the Ornstein-Uhlenbeck process and the linear Langevin process, which we recall shortly and can be found in many books, e.g.~\cite{Ga, Pa, RW}.
Linear stochastic process models have been used for inference in various contexts, e.g.~\cite{HS, RG+, PMPR}.
The point of the present paper is to present general ways in which such models can be used for inference, particularly in systems with many degrees of freedom, to infer their dominant modes of oscillation and to run in streaming mode.

In addition to detecting oscillations in power flow in electricity networks, I envisage the method to be useful in various other contexts, for example detecting soft modes in civil engineering structures, helioseismology, and to study business cycles.  It could also be used for magnetic resonance imaging.

The paper starts by recalling some simple examples of linear stochastic system.  It goes on to review the derivation of the covariance function for a general linear stochastic system.  After a review of probabilistic inference oriented to such models, it comes to the main point, which is how to infer the dominant modes from observations.  A subsequent section describes a way to perform the inference efficiently in real-time.  Then a section on AC electricity networks proposes how to fit them in this scheme.  The paper closes with a discussion.

\section{Simple examples}
\begin{figure}[htbp] 
   \centering
   \includegraphics[width=4in]{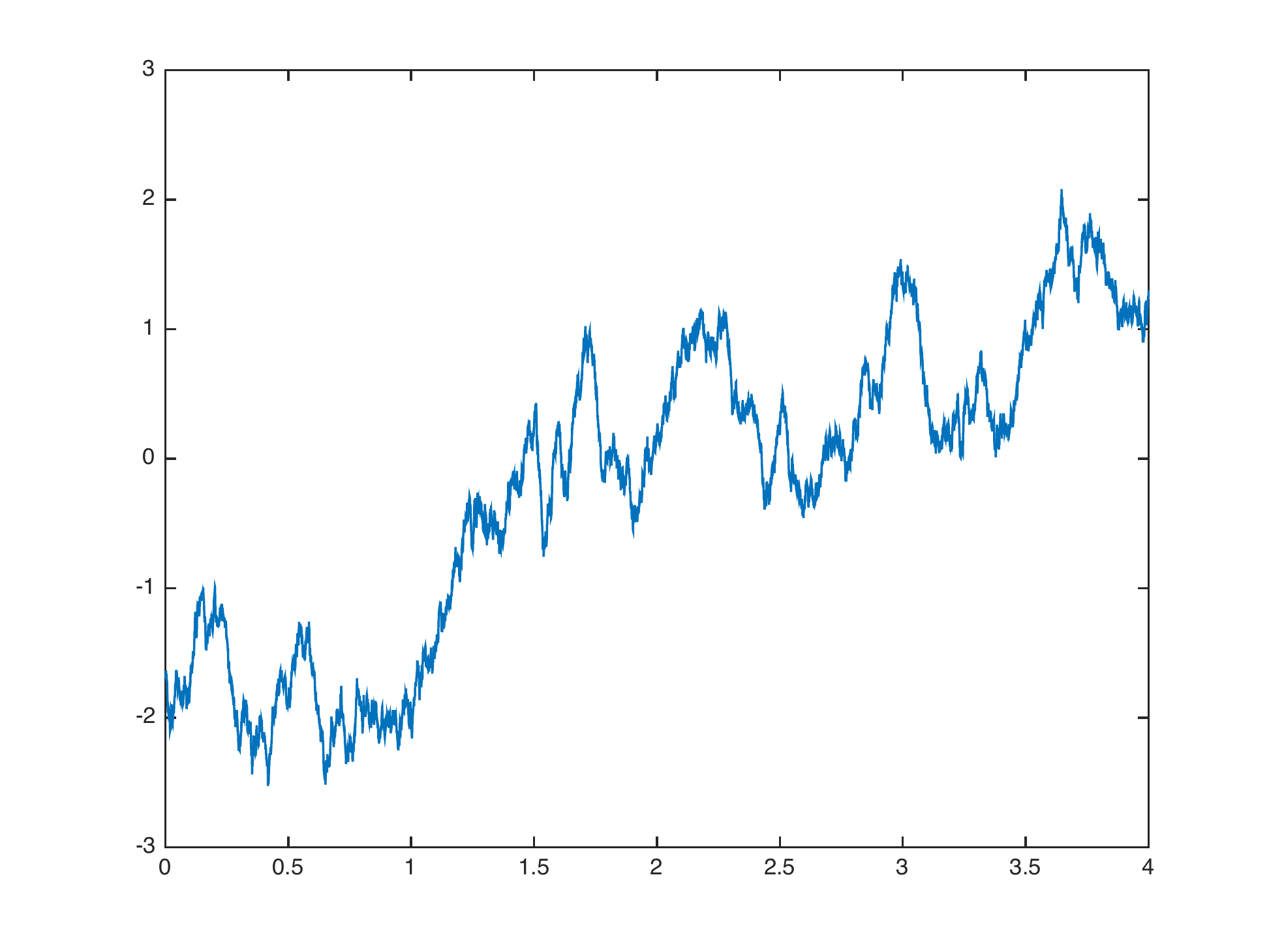} 
   \caption{A sample from the OU process with $\mu=1$, $\sigma = \sqrt{2}$.}
   \label{fig:OUsample}
\end{figure}

The simplest example of asymptotically stable linear system forced by Gaussian noise is the {\em Ornstein-Uhlenbeck (OU) process}:
\begin{equation}
\frac{dx}{dt} = -\mu x + \sigma \xi,
\end{equation}
with $x \in \R$, $\mu >0$, $\sigma > 0$ and $\xi$ Gaussian white noise (which can be considered as a highly degenerate Gaussian process on $\R$ with mean $M(t)=0$ and covariance $C(t,t') = \delta(t-t')$).  Then Duhamel's formula 
\begin{equation}
x(t) = \int_{-\infty}^t e^{-\mu(t-s)} \sigma \xi(s)\ ds
\end{equation}
shows that $x$ is a Gaussian process with mean zero and covariance 
\begin{equation}
C(t,t') = \langle x(t) x(t')\rangle = \frac{\sigma^2}{2\mu} e^{-\mu |t-t'|}.  
\end{equation}
A sample is shown in Figure~\ref{fig:OUsample}.  With probability one, samples are continuous but nowhere differentiable \cite{Ad}.

\begin{figure}[htbp] 
   \centering
   \includegraphics[width=4in]{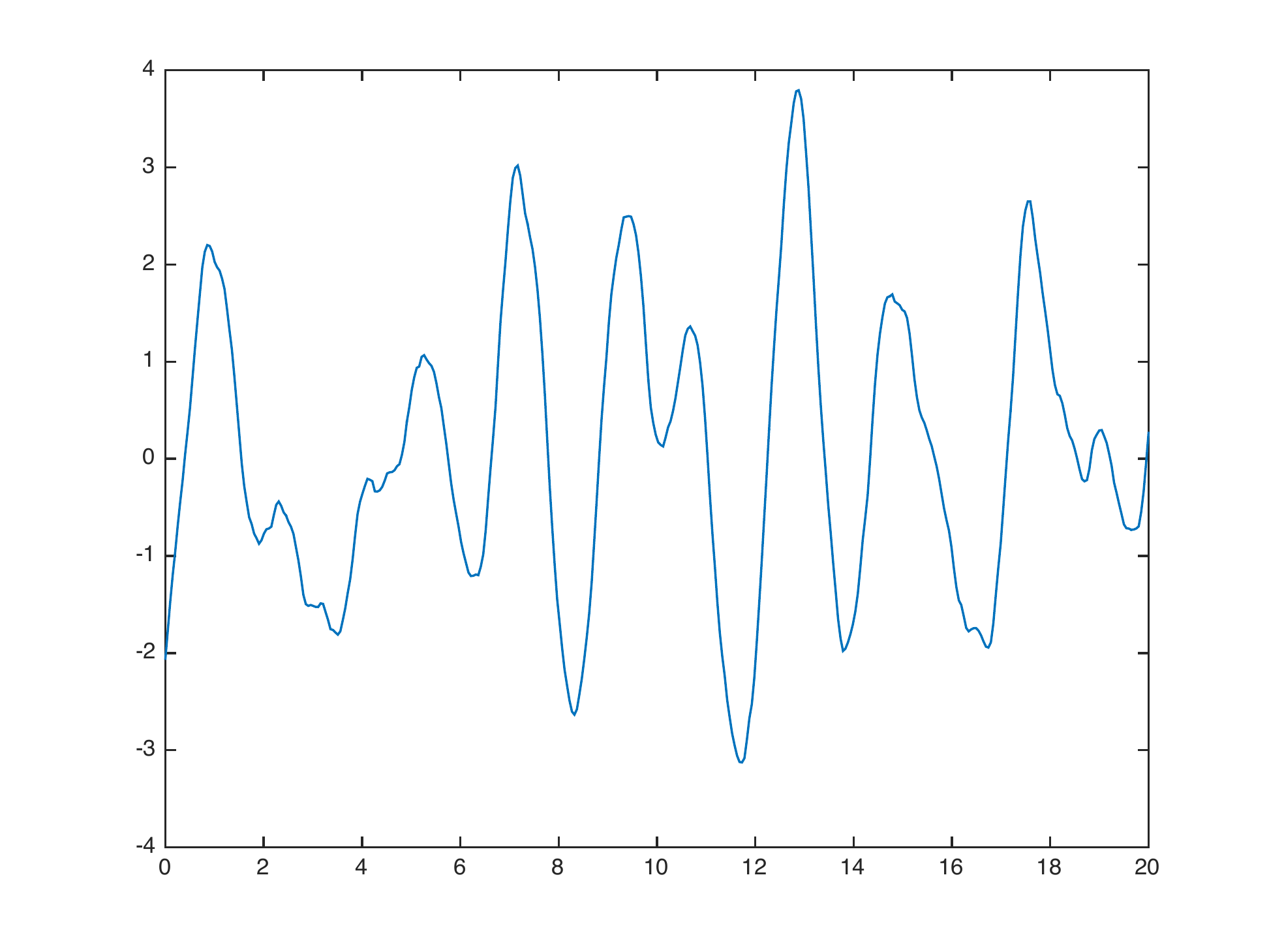} 
   \caption{A sample for the underdamped linear Langevin process with $\sigma^2 = 2\beta k$, $\alpha = 1/e$, $\omega = e$ ($e$ being the base of natural logarithms).}
   \label{fig:ULL}
\end{figure}

Next we consider the {\em linear Langevin process}:
\begin{equation}
m \ddot{x} + \beta \dot{x} + k x = \sigma \xi,
\end{equation}
with $m, \beta, k, \sigma >0$ (cf.~\cite{Pa}).  It follows that $x$ is a GP on $\R$ with mean zero and covariance 
\begin{equation}
C(t,t') = \frac{\sigma^2}{2\beta k} e^{-\alpha |\tau|} (\cos \omega \tau + \frac{\alpha}{\omega} \sin \omega |\tau|), 
\end{equation}
where $\tau = t-t'$, $\alpha = \frac{\beta}{2m}$, $\omega = \frac{1}{m} \sqrt{mk-\beta^2/4}$.  This formula is most appropriate for the underdamped case $\beta^2/4 < mk$.  In the overdamped case $\beta^2/4 > mk$, it is more usefully written as
\begin{equation}
C(t,t') = \frac{\sigma^2}{4\beta k \eps} (\lambda_+ e^{-\lambda_- |\tau|} - \lambda_- e^{-\lambda_+ |\tau|}),
\end{equation}
where $\eps = \frac1m \sqrt{\beta^2/4 - mk}$ and $\lambda_\pm = \alpha \pm \eps$.  In the critically damped case $\beta^2/4=mk$, \begin{equation}
C(t,t') = \frac{\sigma^2}{2\beta k} e^{-\alpha |\tau|} (1 + \alpha |\tau|).  
\end{equation}
A sample for an underdamped case is shown in Figure~\ref{fig:ULL}.  A sample for an overdamped case will appear in Figure~\ref{fig:fou}.  With probability one, solutions of the linear Langevin equation are differentiable but nowhere twice differentiable \cite{Ad}.

The linear Langevin equation can be written as a system of two first-order differential equations.  This can be generalised to the 2D system
\begin{equation}
\dot{x} = Ax + \eta
\end{equation}
where $x \in \R^2$, $A$ a $2\times 2$ matrix with $\tr\ A< 0$, $\det A> 0$, and $\eta \in \R^2$ with $\eta$ being 2D Gaussian white noise with (psd) covariance matrix $K$ (i.e.~$\langle \eta_i(s)\eta_j(t)\rangle = K_{ij}\delta(t-s)$).  Then $x$ is a GP on $\{1,2\} \times \R$, the first factor indicating the component of $x$ (for which we use subscript notation).  It has zero mean.  Its covariance function, which we write as a matrix function on $\R^2$ is
\begin{equation}
C(t,t') = \begin{cases}
\Sigma e^{A^T(t-t')} & \text{ for } t>t' \\
e^{A(t'-t)} \Sigma & \text{ for } t'>t , 
\end{cases}
\end{equation}
where 
\begin{equation}
\Sigma = \int_0^\infty d\sigma\ e^{A\sigma} K e^{A^T \sigma}.
\end{equation}

Taking one component of the general 2D system produces a family of covariance functions that we advocate for purposes such as deciding if a system is under- or over-damped \cite{MP}.

\section{General linear stochastic system}

In this section we review the calculation of the mean and covariance functions for an asymptotically stable continuous-time forced linear system of arbitrary dimension.  Initially, we allow the system to be non-autonomous and we do not restrict the forcing to be Gaussian.  Thus we consider
\begin{equation}
\dot{x}(t) = A(t) x(t) + \eta(t)
\label{eq:linsys}
\end{equation}
with $x, \eta \in \R^n$.
The asymptotic stability assumption implies that the response $x$ to forcing $\eta$ can be written as
\begin{equation}
x(t) = \int_{-\infty}^t H(t,t')\eta(t')\ dt',
\end{equation}
with $H$ the impulse response (matrix-valued Green function), i.e.~the matrix solution of
\begin{equation}
\frac{\partial H}{\partial t} = A(t) H(t,t')
\end{equation}
for $t>t'$ with $H(t'+,t') = I$.  Note that for any $t<t'<t''$,
\begin{equation}
H(t,t'') = H(t,t')H(t',t'').
\label{eq:composition}
\end{equation}

If $\eta$ is a Gaussian process on $\{1,\ldots n\} \times \R$ with mean function $M^\eta$ and covariance function $C^\eta$ (so $C^\eta(s,t) = \langle\eta(s)\eta^T(t)\rangle$), then $x$ is a GP on the same set, with mean function 
\begin{equation}
M^x(t) = \int_{-\infty}^t H(t,t')M^\eta(t')\ dt'
\end{equation}
and covariance function
\begin{equation}
C^x(s,t) = \int_{-\infty}^s\ ds'  \int_{-\infty}^t\ dt'\ H(s,s')C^\eta(s',t')H^T(t,t').
\end{equation}

If the system is autonomous then $H(s,s')$ is a matrix-function $h(\sigma) = e^{A\sigma}$ of just one variable $\sigma = s-s'$.
If the forcing is stationary then $M^\eta$ is constant and $C^\eta(s,t)$ is a matrix-function $k(\tau)$ of $\tau = t-s$ and $k(-\tau) = k(\tau)^T$.  
So assuming both and changing variables to $\sigma$ and $\tau' = t'-s'$,
\begin{eqnarray}
M^x &=& \left(\int_0^\infty h(\sigma)\ d\sigma\right) M^\eta = -A^{-1} M^\eta, \\
C^x(s,t) &=& \int_0^\infty d\sigma \int_{-\infty}^{\tau+\sigma} d\tau' \ h(\sigma)k(\tau')h^T(\tau+\sigma-\tau'). \label{eq:covt}
\end{eqnarray}

Now we specialise further to forcing of zero-mean and white, i.e.~$k(\tau) = K \delta(\tau)$ for some psd symmetric matrix $K$.  
A common way to write this is $\eta(t) = B \xi(t)$ for $\xi$ a vector of independent unit Gaussian white noises and a matrix $B$; then $K=BB^T$.  But $B$ can be replaced by $BO$ for any orthogonal matrix without changing the probability distribution for $\eta$, so this description contains useless redundancy and it is better to specify the noise $\eta$ by just its covariance matrix $K$.

In this case, $x$ has zero mean and
\begin{equation}
C^x(\tau) = \int_0^\infty d\sigma\ h(\sigma) K h^T(\tau+\sigma) \mbox{ for } \tau > 0.
\end{equation}
For $\tau < 0$, $C^x(\tau) = C^x(-\tau)^T$.
Using $h(\tau+\sigma) = h(\sigma)h(\tau)$ for $\sigma, \tau > 0$ (a special case of (\ref{eq:composition})), this boils down to
\begin{equation}
C^x(\tau) = \begin{cases}
\Sigma  h^T(\tau) & \text{ for } \tau > 0 \\
h(-\tau) \Sigma & \text{ for } \tau < 0,
\end{cases} \label{eq:cov}
\end{equation}
where the symmetric matrix
\begin{equation}
\Sigma = \int_0^\infty d\sigma\ h(\sigma) K h^T(\sigma),
\label{eq:Sigma}
\end{equation}
giving the result that the covariance of the response of an asymptotically stable autonomous linear system to Gaussian white noise is a matrix multiple of the transpose of the impulse response function (for $\tau>0$) (e.g.~p.105 of \cite{Ga}).  

Note that $\Sigma$ satisfies the Sylvester equation \cite{Ga} (actually its special case due to Lyapunov):
\begin{equation}
A\Sigma + \Sigma A^T = -K.
\label{eq:Lyap}
\end{equation}
The theory of Sylvester equations (e.g.~\cite{BR}) shows that it has a unique solution for $\Sigma$ because $A$ has been assumed to have all its spectrum in the open left half plane, so there are no pairs $(\lambda_i,\lambda_j)$ of eigenvalues for $A$ and $A^T$ that sum to zero.
An interesting approach using (\ref{eq:Lyap}) to infer $A$ adn $K$ from $\Sigma$ in the AC electricity context is presented in \cite{WBT}, where the model is called a vector OU process.

\section{Inference}
\label{sec:inf}
We review the standard approach to inference of a dynamical system from observations.  In our case, the system is modelled by
\begin{equation}
\dot{x}=Ax + \eta,
\end{equation}
with $\langle \eta(s)\eta^T(t)\rangle = K\delta(t-s)$.  This needs augmenting by a model for the observations, e.g.~vectors 
\begin{equation}
y_i = C_i x(t_i) + \eps_i
\end{equation}
for some known sequence of times $t_i$, known observation matrices $C_i$, and unknown measurement error $\eps_i$ which could be assumed to be independent zero-mean Gaussian vectors with unknown covariance matrices $k_i$, but with $k_i=k_j$ if $C_i = C_j$; the idea is that the matrices $C_i$ specify which components (or combinations of components) of $x$ are measured.

Then the parameters of the model are the matrix elements of $A$ and $K$ and of the $k_i$ (though we are less interested in the latter, so let us ignore them).  If $x$ has dimension $L$, the parameters form a continuous space ${\mathcal P}$ of dimension $L^2 + L(L+1)/2 = L(3L+1)/2$ (though this may be reduced significantly if the system has known structure).  Let us denote the parameters in short by a vector $\mu$.

Previous knowledge about the system is encoded into a prior probability density $P_-(\mu)$ for the parameters.  After the observations $Y$, a posterior probability density $P_+$ is computed for the parameters by Bayes' rule:
\begin{equation}
P_+(\mu|Y) = P_s(Y|\mu) P_-(\mu)/Z(Y)
\end{equation}
where $P_s$ is the probability density for the observations given the parameters, which is specified by the model, and $Z = \int P_s(Y|\mu) P_-(\mu)\ d\mu$ is a normalisation factor.

If enough observations have been taken (depending on how tight the prior $P_-$ was), then $P_+$ will be tightly peaked around some value $\hat{\mu}$ of the parameter vector.  The maximum posterior likelihood value is the $\hat{\mu}$ that maximises $P_+(\mu|Y)$ (assuming it is unique).  Although the functional form for $P_+$ is not in general computable, numerical algorithms like gradient ascent can search for $\hat{\mu}$.  They can also compute a quadratic approximation to $P_+$ around $\hat{\mu}$ to give an idea of the posterior uncertainty in the inference.

Once a best fit to $A$ and $K$ has been obtained, one could compute the modes of $A$ (frequency, damping, shape) and find the covariance matrix between their amplitudes induced by the forcing $K$.

The main point of this paper, however, is that the above is overkill.  If we are interested only in the dominant modes we can infer them without inferring $A$ and $K$.  We will show that to infer $N$ modes (counting complex ones twice) from $M$ observation components requires a parameter space of dimension only $(N+1)(M+N/2)$.  This is likely to be much less than the dimension $L(3L+1)/2$ of the space of $A$ and $K$ above, because both $N$ and $M$ are smaller than $L$.

A second point of the paper is that the inference can be run in streaming mode, with the posterior probability and the maximum posterior likelihood estimate of parameters being updated as each new observation arrives.  It can be done efficiently, with each new observation requiring the same time to process regardless of how many previous observations have been made, whereas for a general GP the computation time to infer from $J$ observations scales like $J^3$ and the time to add one new observation scales like $J^2$.

\section{Fitting dominant modes}

Rather than attempting to fit the whole matrix $A$ (and $K$) to observations, we propose to fit just the dominant modes.

The system matrix $A$ can always be put into a block-diagonal form $D$, i.e.~$A = B D B^{-1}$ for some invertible matrix $B$, thus so can the impulse response matrix function $h(t) = B e^{Dt} B^{-1}$.  So the covariance function for the response can be written (for $\tau>0$) as
\begin{equation}
C^x(\tau) = B S e^{D^T\tau} B^{T},
\label{eq:resp}
\end{equation}
with $S = \int_0^\infty d\sigma\ e^{D\sigma} B^{-1}KB^{-T} e^{D^T\sigma}$.
In particular, if $A$ has simple eigenvalues then it can be put into such a form with the diagonal blocks of $D$ being one- or two-dimensional (we prefer to avoid complex coordinate changes).  Each 1D block is a real (negative) eigenvalue $-\lambda$ of $A$.  Each 2D block can be put into the form
\begin{equation}
\left[\begin{array}{cc} -\alpha & -\omega \\
\omega & -\alpha \end{array} \right]
\end{equation}
for complex conjugate pair of eigenvalues $-\alpha \pm i \omega$.
Thus the diagonal blocks of $e^{D^T\tau}$ are $e^{-\lambda \tau}$ for a 1D block and 
\begin{equation}
E^T(\tau) = e^{-\alpha \tau} \left[ \begin{array}{cc} \cos \omega \tau & \sin \omega \tau \\
-\sin \omega \tau & \cos \omega \tau \end{array} \right]
\end{equation}
for a 2D block.

So we make the ansatz that (for $\tau>0$)
\begin{equation}
\label{eq:ansatz}
{C^x}_{ij}(\tau) = \sum_{m,n} B_{im} S_{mn} E^T_n(\tau) B_{jn}
\end{equation}
for some reduced set of modes $m,n$, with $E^T_n$ of the form $e^{-\lambda_n \tau}$ for a real mode $n$ and $E^T$ above for a complex mode.  For a real mode $m$, $B_{im}$ is a column vector indexed by components $i$ of $x$.  For a complex mode $m$, $B_{im}$ is a pair of column vectors.  The columns of $B$ specify the mode shapes.
$S$ is a psd covariance matrix for the modes, whose diagonal elements give the squared amplitudes for each mode and off-diagonal elements specify covariances between the modes.

There are some redundancies in this specification.  Firstly, the order in which the modes are labelled is irrelevant.  One could eliminate this freedom by choosing to label them in order of size of $\lambda$ and $\alpha$.
Secondly, each mode vector can be scaled by an arbitrary non-zero scalar (real for a real mode, complex for a complex mode), subject to scaling $S$ by the inverse square root.  One could eliminate this freedom by for each mode $n$ selecting a ``large'' component $i_n$ and setting $B_{i_n n} = +1$ for a real mode, $[+1,0]$ for a complex mode.
But as one explores parameter space, one may need to change these choices.

Also, we need to enforce that $S$ is psd.  One way to achieve this is to write $S=e^R$ for $R$ symmetric.  There are efficient algorithms for exponentiating matrices.  Another is to write $S=LL^T$ with $L$ lower triangular (in some chosen order on modes), but the diagonal elements of $L$ should be chosen non-negative to remove another redundancy of sign.  Such a Cholesky decomposition is a common step for efficient matrix computations so could essentially come for free.

It might be that a complex mode is close to transition to a pair of real modes, or vice versa.  To allow parameter search in a uniform way near such a transition, it would be better to generalise complex modes to allow pairs of real modes, as in \cite{MP}, but we leave incorporating that refinement to the future.

The number of modes to attempt to fit can be decided by Bayesian model comparison \cite{M2}.  This is an extension of maximum posterior likelihood search to a setting with two or more models $M_j$, which each have their own continuous parameter spaces ${\mathcal P}_j$.  For each model $M_j$, one can compute the posterior probability density $P_+(\mu|Y,M_j)$ for $\mu \in {\mathcal P}_j$.  By various methods, e.g.~\cite{AK+}, one can also compute the normalisation constant $Z(Y|M_j)$, called {\em Bayes' factor} for the model.
Then given prior probabilities $P_-(M_j)$ for the models (which can be taken the same if one is agnostic about which model is best), one applies Bayes' rule again to obtain posterior probabilities
\begin{equation}
P_+(M_j|Y) = Z(Y|M_j)P_-(M_j)/Z(Y)
\end{equation}
where $Z(Y)$ is a normalisation factor again, depending on $Y$ and the chosen set of models, but is not required for what follows.
This formula can be used to decide which model is the best explanation of the observations and to keep track of near-competitors.
For each model $M_j$, the method of Section~\ref{sec:inf} determines best-fit parameters $\hat{u}_j \in {\mathcal P}_j$.
In our case the different models correspond to the numbers $N_R$ and $N_C$ of real and complex modes to fit.  The idea is that even though a better fit is achievable with more modes, the increased dimension of parameter space might not justify using it (Occam's razor).

\section{Streaming data}

In many circumstances it would be preferable to run the inference of modes in real time rather than batch, and efficiently.  There are papers on real-time inference with GPs, e.g.~\cite{RR, RG+, HS, BNT}, in particular using the Kalman filter, but I didn't find one that goes as far as I want.  

I propose that a good way to infer the state of an autonomous continuous-time linear system forced by white noise from real-time observations is the following version of the Kalman filter.  We denote the state of the system at time $t\in \R$ by $x(t) \in \R^n$ and we suppose it evolves by
\begin{equation}
\dot{x} = Ax + \eta,
\end{equation}
with $\eta \in \R^n$ Gaussian white noise of covariance matrix $C^\eta$.
We suppose observations are taken at an increasing sequence of times $t_i$.  In contrast to claims in some of the literature, they do not need to be equally spaced and one can observe different components of $x$ at different times.  So we let the observations be
\begin{equation}
y_i = Z_i x_i + m_i + \xi_i
\end{equation}
where $y_i, m_i,\xi_i \in \R^{d_i}$, $x_i = x(t_i)$ and $\xi_i$ is a zero-mean Gaussian measurement noise which we suppose independent for different $i$.

Then for a sequence of vectors $x_i$ at the times $t_i$, use the notation $x_{i|i-1} = \langle x_i | y_{i-1},\ldots y_1\rangle$ and $x_{i|i} = \langle x_i| y_{i},\ldots y_1\rangle$.  Similarly define $y_{i|i-1} = \langle y_i | y_{i-1},\ldots y_1 \rangle$.  Let 
\begin{equation}
P_{i|i-1} = \langle (x_i-x_{i|i-1})(x_i-x_{i|i-1})^T \rangle 
\end{equation}
and similarly $P_{i|i} = \langle (x_i-x_{i|i})(x_i-x_{i|i})^T \rangle$.  Write $\tau_i = t_i-t_{i-1}$.
As a consequence of the Duhamel formula
\begin{equation}
x_i = e^{A\tau_i} x_{i-1} + \int_{t_{i-1}}^{t_i} e^{A(t_i-t)}\eta(t)\ dt,
\end{equation}
we obtain
\begin{equation}
x_{i|i-1} = e^{A\tau_i} x_{i-1|i-1}
\end{equation}
and
\begin{equation}
P_{i|i-1} = e^{A\tau_i} P_{i-1|i-1}e^{A^T\tau_i} + G_i ,
\end{equation}
with
\begin{equation}
G_i = \int_{t_{i-1}}^{t_i} e^{A(t_i-t)}C^\eta e^{A^T(t_i-t)}\ dt.
\end{equation}
Also
\begin{equation}
y_{i|i-1} = Z_i x_{i|i-1} + m_i .
\end{equation}
Let 
\begin{eqnarray}
v_i &=& y_i - y_{i|i-1} , \\
F_i &=& \langle v_i v_i^T | y_{i-1},\ldots y_1\rangle . \label{eq:Fi}
\end{eqnarray}
Then 
\begin{equation}
F_i = Z_i P_{i|i-1} Z_i^T + H_i ,
\end{equation}
where $H_i$ is the covariance matrix of $\xi_i$.
Finally, by conditioning on $y_i$, we obtain
\begin{eqnarray}
x_{i|i} &=& x_{i|i-1} + K_i v_i , \\
P_{i|i} &=& (I - K_i Z_i) P_{i|i-1} ,
\end{eqnarray}
where the ``Kalman gain matrix"
\begin{equation}
K_i = P_{i|i-1}Z_i^T F_i^{-1}.
\end{equation}

The standard use of these equations is to provide an estimate $x_{i|i}$ of the state $x_i$.  But they can also be used to provide the likelihood for parameters of the model, given the observations, and this is our primary goal.  To see this, 
the likelihoods $f$ for the observations given the parameter values satisfy
\begin{equation}
f(y_i,\ldots y_1) = f(y_i|y_{i-1},\ldots y_1) f(y_{i-1},\ldots y_1).
\end{equation}
So from (\ref{eq:Fi}), the {\em evidence} for the model, defined to be the log-likelihood of the observations as a function of the parameters, updates by
\begin{equation}
L_i = \log f(y_i,\ldots y_1) = L_{i-1} - \frac12 (v_i^T F_i^{-1}v_i+  \log \det F_i+ d_i \log 2\pi),
\end{equation}
where we recall that $d_i$ is the dimension of the observation vector $y_i$ at time $t_i$.
This provides the total evidence for the given parameters ($A,C^\eta,Z_i,m_i,H_i$), starting from the initial time.  
Despite the fact that for a general GP it takes time $O(N^3)$ to compute the likelihood from $N$ observations, the class of linear stochastic processes with the above algorithm takes equal time per observation, allowing the computation to be done in real-time.

One can similarly (albeit messily) work out how to update the derivative of $L_i$ with respect to the parameters; use that the derivative $(\log\det F)' = \tr (F^{-1}F')$ where prime denotes derivative with respect to any parameter.
Thus one can make gradient steps to improve the estimate of the maximum-likelihood parameters.

To adapt to the case where the parameters may in reality be slowly varying it is better not to maximise the evidence for the whole time-interval of observation but instead to maximise an exponentially weighted sum of the gains in evidence, allowing one to forget past evidence because it is likely to become irrelevant.  Choose a rate constant $\lambda$ for forgetting past evidence.  The evidence gained at time $t_{i}$ relative to $t_{i-1}$ is
\begin{equation}
\eps_i = - \frac12 (v_i^T F_i^{-1}v_i+  \log \det F_i+ d_i \log 2\pi) .
\end{equation}
An appropriate notion of the weighted sum of gains, that I call {\em discounted evidence rate}, is 
\begin{equation}
\tilde{L}_i = \sum_{j=1}^i e^{-\lambda(t_i-t_j)} \eps_j, 
\end{equation}
and it updates by
\begin{equation}
\tilde{L}_i = e^{-\lambda\tau_i} \tilde{L}_{i-1} + \eps_i.
\end{equation}
Again, derivative information can be updated and gradient steps made to track maximum likelihood parameters.

In principle, one can begin by specifying a prior probability density on the parameter space but its effect on the discounted evidence rate will go to zero exponentially in the time since the start.

If one wants to allow the number of modes to vary then one needs to do Bayesian model comparison again, by running several different models alongside each other and computing their Bayes' factors.

\section{AC electricity networks}
\label{sec:ACelec}
We turn now to the motivating application.

The dynamics of an AC (alternating current) electricity network can be modelled approximately by a connected graph with a node for each rotating machine (synchronous generator or motor) \cite{MBB} (this leaves open the question of how to model DC/AC convertors, such as at wind farms, solar photovoltaic farms and DC interconnector terminals).  Let $N$ be the number of nodes.  
As described in \cite{Rog} (another useful reference is \cite{An}), one can model an AC network at various levels of complexity.  If one ignores aspects like the dynamics of the voltages\footnote{This is relatively easy to incorporate, e.g.~\cite{TBP}, but a full treatment would require including voltage control, power system stabilisers, and excitor control}, 3-phase imbalances, reactive power control and harmonics, the state can be specified by a phase $\phi_l$ and frequency\footnote{As $\phi_l$ is in radians it might be better to denote $f_l$ by $\omega_l$, but I am already using $\omega$ for mode frequencies} $f_l = \dot{\phi}_l$ at each node $l$,
and dynamics for the vector $f$ of frequencies and phases $\phi$ are given by balancing power (cf.~(1) of \cite{SMH} or (17) of \cite{SM}):
\begin{eqnarray}
I_l f_l \dot{f}_l &=& p_l - \Gamma_l f_l^2  -\sum_{l'} V_l V_{l'} \left(B_{l l'} \sin(\phi_l - \phi_{l'}) + G_{l l'} \cos(\phi_l - \phi_{l'}) \right) \label{eq:ACpower} \\
\dot{\phi}_l &=& f_l \nonumber
\end{eqnarray}
where $I_l$ is an inertia, $\Gamma_l$ a damping constant, $V_l$ is the amplitude of the voltage at $l$, $B_{l l'}$ is a symmetric matrix of ideal admittances of the line between $l$ and $l'$ ($B_{ll}=0$), $G_{l l'}$ is a symmetric psd matrix of conductances of the line between $l$ and $l'$ (which produces transmission losses) including self-conductances,
and $p$ is a vector of power imbalances (generation minus consumption), which is to be regarded as an external stochastic process (e.g.~people switching loads on and off, wind farms producing varying power).  For the moment, think of $p$ as fixed.
For an example of more detailed modelling, see \cite{JK}.

Note that it is common in the electrical engineering literature to partially linearise (\ref{eq:ACpower}) about a reference frequency $f_0$ (usually $100\pi$ or $120\pi$ sec$^{-1}$) by writing $\omega_l = f_l-f_0$, $\delta_l = \phi_l - f_0 t$, and replace $I_lf-l\dot{f}_l$ by $M_l \dot{\omega}_l$ with $M_l = I_l f_0$ (which is often called an inertia again) and $\Gamma_l f_l^2$ by $D\omega_l$ with $D=2\Gamma_l f_0$.  I shall completely linearise later in this section, but for the present retain the fully nonlinear form (\ref{eq:ACpower}) for discussion of its global phase symmetry and its equilibria.

The system has the special feature of global phase-rotation invariance: if one adds the same constant to all the phases then the dynamics produce the same trajectory but with the constant added.  One can quotient by this symmetry group, which we denote by $S$.\footnote{In reality, the system operator is required to keep the phases within some interval (of about 100 cycles) around that for a reference rotor at the nominal frequency, so they exert changes to $p$ to achieve this, thereby breaking the phase rotation invariance, but we will ignore that.}  For example, choose a root node $o$ and a spanning tree in the graph, orient its edges $e$ away from $o$ (other choices are alright but this is to make a definite choice), and let $\Delta_e = \phi_{l'}-\phi_l$ for each edge $e = l l'$ in the spanning tree; there are $N-1$ of these, and we denote the vector of phase differences by $\Delta$.  Then the phase difference between any two nodes can be expressed as a signed sum of the $\Delta_e$, and the equations $\dot{\phi}_l = f_l$ can be replaced by $\dot{\Delta}_e = f_{l'}-f_l$.

The quotient system has a manifold of equilibria in the space of all power imbalance vectors $p$, frequency vectors $f$ and phase difference vectors $\Delta$.  For an equilibrium (mod $S$), each node has the same frequency and the phase differences are constant.   The manifold of equilibria is a graph over the space of common frequency $F \in \R$ and phase differences $\Delta \in (\R/2\pi\Z)^{N-1}$:
\begin{equation}
p_l = \Gamma_l F^2  + \sum_{l'} V_l V_{l'} \left (B_{l l'} \sin(\phi_l - \phi_{l'}) + G_{l l'} \cos(\phi_l - \phi_{l'}) \right). 
\end{equation}
Let us restrict attention to the part with $F$ near a nominal reference frequency $F_0$ (50Hz in Europe, which means $F_0 = 100 \pi$ in radians/sec) and $p_l$ near $\Gamma_l F_0^2 + \sum_{l'} V_l V_{l'} G_{ll'}$.  Then there is a stable equilibrium with all phase differences between linked nodes near zero.
The stability can be established by the energy method used in \cite{TBP}, modified to include the conductance matrix $G$ and ignore the voltage dynamics.  It should be noted, however, that inclusion of governors or power system stabilisers in the model can destabilise the equilibrium and produce oscillations \cite{Rog}, presumably by a Hopf bifurcation.  The method of the present paper is not well adapted to detecting autonomous oscillations as opposed to damped ones forced by noise.

Suppose the system is near the stable equilibrium for some $p$.  As $p$ moves in time, the response roughly follows it on the manifold of equilibria, but deviations from equilibrium are in general excited and these would relax back to equilibrium if $p$ were to stop moving.  For small movements of $p$ about a mean imbalance vector $P$ with corresponding stable equilibrium $(F, \Delta)$, it is appropriate to linearise the system.  A reference for small-signal stability in power systems is \cite{GPV}.
Write $\delta f_l$, $\delta \Delta_e$, $\delta p_l$ for the deviations of $f_l$, $\Delta_e$ and $p_l$ from the equilibrium.  Write
\begin{equation}
M_l = I_l F, \quad \gamma_l = 2\Gamma_l F,
\end{equation}
\begin{equation}
T_{ll'} = V_l V_{l'} (B_{ll'}\cos(\Phi_l-\Phi_{l'})-G_{ll'}\sin(\Phi_l-\Phi_{l'})).
\end{equation}  
Then
\begin{eqnarray}
M_l \dot{\delta f}_l &=& \delta p_l - \gamma_l \delta f_l - \sum_{l'} T_{ll'} (\delta \phi_l - \delta \phi_{l'}) \\
\dot{\delta \Delta}_e &=& \delta f_{l'}-\delta f_l \nonumber
\end{eqnarray}
for $e = l l'$.  Write this as
\begin{equation}
\dot{x} = Ax + C \delta p
\label{eq:x}
\end{equation}
with $x = \left[ \begin{array}{c} \delta f \\ \delta \Delta \end{array} \right]$ and $C = \left[ \begin{array}{c} \diag M_l^{-1} \\ 0 \end{array} \right]$.

We choose to model the dynamics of the power imbalances by
\begin{equation}
\dot{\delta p} = - J \delta p + \sigma \xi
\label{eq:deltap}
\end{equation}
for some matrix $J$ (with $-J$ asymptotically stable) and (multidimensional) Gaussian white noise $\sigma \xi$ with covariance matrix $K = \sigma \sigma^T$ (later, $J$, $P$, $T$ and $K$ may vary slowly in time).  This is a somewhat crude representation, but captures the idea that $p$ has random increments and reversion to a mean.  There is evidence that load distribution is close to Gaussian, e.g.~fig.14 of \cite{TT+}, which is consistent with this model, though that data says nothing about the temporal correlations.  It is common to neglect temporal correlations of the power imbalance, e.g.~\cite{WBT}, but there are automated and human responses to power imbalance which have a filtering effect.
One might argue that National Grid's balancing actions are based more on the deviations of the average frequency and phase differences from nominal than the power imbalances, but on the manifold of equilibria these are equivalent.

The resulting system (\ref{eq:x}, \ref{eq:deltap}) for $(x,\delta p)$ is of the form (\ref{eq:linsys}), but it has a skew-product structure that we should exploit, namely $\dot{\delta p}$ does not depend on $x$ (also the $x$-dynamics has structure in that it is only the frequencies that see $\delta p$ directly).  In reality, perhaps $\dot{\delta p}$ does depend a little on $x$, e.g.~National Grid balancing operations and frequency-sensitive generators and loads, but let us continue with this model.  One way to exploit the skew-product structure is to derive the covariance function for $\delta p$ using (\ref{eq:cov}) and then insert this into the formula (\ref{eq:covt}) for the covariance function of $x$, but it leads to an integration whose treatment is not simple.  Alternatively, we can apply (\ref{eq:cov}) to the joint system (\ref{eq:x}, \ref{eq:deltap}), exploit the skew-product form of the impulse response, and take the $xx$-block of the covariance function.  I chose the latter approach, subject to the simplifying but generic assumption of simple eigenvalues for the full system.

The impulse response of (\ref{eq:deltap}) can be written in matrix exponential notation as $\delta p(t) = e^{-Jt}$.  Similarly, the impulse response of (\ref{eq:x}) can be written as $x(t) = e^{At}$.  To compute the response of $x$ to an impulse on $\dot{p}$, it is convenient to assume that $A$ and $-J$ have no eigenvalues in common, as is generically the case.  Then there exists a unique solution $E$ to another Sylvester equation 
\begin{equation}
AE+EJ = C,
\label{eq:E}
\end{equation}
and defining $y=x+Ep$ we see that $\dot{y} = Ay + E\xi$.
So the response of $y$ to an impulse on $\dot{p}$ is $e^{At}E$.
It follows that the response of $x = y-Ep$ to an impulse on $\dot{p}$ is 
\begin{equation}
x(t) = h_{xp}(t) : = e^{At}E-E e^{-Jt}.
\end{equation}
Note that using (\ref{eq:E}), the time-derivative of $h_{xp}$ at $t=0$ is just $C$.
Thus the impulse response of the full system has the block form
\begin{equation}
h(t) = \left[ \begin{array}{cc}
e^{-Jt} & 0 \\
e^{At} E - E e^{-Jt} & e^{At} \end{array} \right] .
\label{eq:h}
\end{equation}
Then the stationary covariance matrix $\Sigma$ (\ref{eq:Sigma}) of the joint process has the block form
\begin{eqnarray}
\Sigma &=& \int_0^\infty  h(\sigma) \left[ \begin{array}{cc} K & 0 \\ 0 & 0 \end{array} \right] h^T(\sigma)\ d\sigma \nonumber\\
&=& \int_0^\infty 
\left[ \begin{array}{cc} e^{-J\sigma} K e^{-J^T \sigma} & e^{-J\sigma} K h_{xp}^T(\sigma) \\
h_{xp}(\sigma) K e^{-J^T\sigma} & h_{xp}(\sigma) K h_{xp}^T(\sigma) \end{array} \right] d\sigma .
\end{eqnarray}
It follows from (\ref{eq:cov}) that (for $\tau > 0$)
\begin{eqnarray}
&C^{x}(\tau) = \Sigma_{xp} h_{xp}^T(\tau) + \Sigma_{xx} e^{A^T \tau} \\
&= \left(\int_0^\infty  h_{xp}(\sigma) K E^T e^{A^T \sigma} d\sigma \right) e^{A^T \tau}
- \left (\int_0^\infty h_{xp}(\sigma) K e^{-J^T\sigma} d\sigma \right) e^{-J^T\tau} E^T . \nonumber
\end{eqnarray}

Thus the covariance of $x=(\delta f,\delta \Delta)$ is a linear combination of functions from the impulse response of $x$ to $\dot{x}$ and of $p$ to $\dot{p}$.\footnote{In the case of common eigenvalues $\lambda$ to $-J$ and $A$ there would in general also be terms of the form $P(\tau) e^{\lambda \tau}$ with $P$ a polynomial of degree higher than those which might already result from multiplicity in $-J$ or $A$.} 

So now we can fit observations of $(f, \Delta)$ at as many locations as available (say, $k$) and as a function of time $t$ to an autonomous GP with mean function of the form $(F {\bf1}, \bar{\Delta})$ for some $F \in \R$ and $\bar{\Delta} \in \R^{k-1}$ and covariance function of the form (\ref{eq:resp}).  We make the obvious step of shrinking the spanning tree to one for just the observed nodes.

So the proposal is to fit an autonomous GP with mean function $(F,\bar{\Delta})$ and covariance function of the form (\ref{eq:resp}) to observations $(f_l,\Delta_e)$ as functions of time $t$, but with $B$ truncated to have only a small number of columns, i.e.~(\ref{eq:ansatz}).  
The observations can be deduced from phasor measurement units (PMUs), which measure (among many things) the (voltage) phase relative to a notional 50Hz reference and the instantaneous frequency at their location.
For $N_R$ real modes and $N_C$ complex modes and $M$ observation components ($f_l$ for each PMU $l$ and $\Delta_e$ for the voltage phase difference along each edge $e$ in the spanning tree of the PMUs, so $M = 2k-1$ where $k$ is the number of PMUs), the parameter space consists of $N_R$ decay rates $\lambda_n$ for the real modes, $N_C$ frequencies $\omega_m$ and decay rates $\alpha_m$ for the complex modes, $N_R$ vectors $B_{in}$ of length $M$ for the real mode shapes normalised to have one component $+1$, $N_C$ pairs of vectors $B_{im}$ of length $M$ for the complex modes normalised to have one component $(+1,0)$, $N(N+1)/2$ coefficients of the mode correlation matrix $S$ (symmetric), where $N = N_R + 2 N_C$, 1 mean frequency $F$ and $k-1$ mean phase differences along the edges of the spanning tree.  This makes a total dimension $N(2k+\frac{N-1}{2})+k$ of parameter space.  This is slightly less than the dimension stated in Section~\ref{sec:inf}, because for the AC electricity system it is automatic that the time-mean frequencies at all PMUs are the same.  If one adds $k-1$, one obtains the dimension claimed there.  
If one desires to fit many modes, this dimension could be quite large, but it is still much smaller than the dimension of the parameter space for the whole system.

As an example, if there are $k=10$ PMUs and one wishes to fit 2 real modes and 1 complex mode then $N=4$ and the parameter space has dimension 53.  One might say one is not interested in real modes but they are probably the dominant ones and to detect a complex mode one needs to fit the dominant behaviour too.

There is the question of how many modes to allow, both real and complex.  This can be decided by the Bayesian comparison method already mentioned.

\begin{figure}[htbp] 
   \centering
   \includegraphics[width=5in]{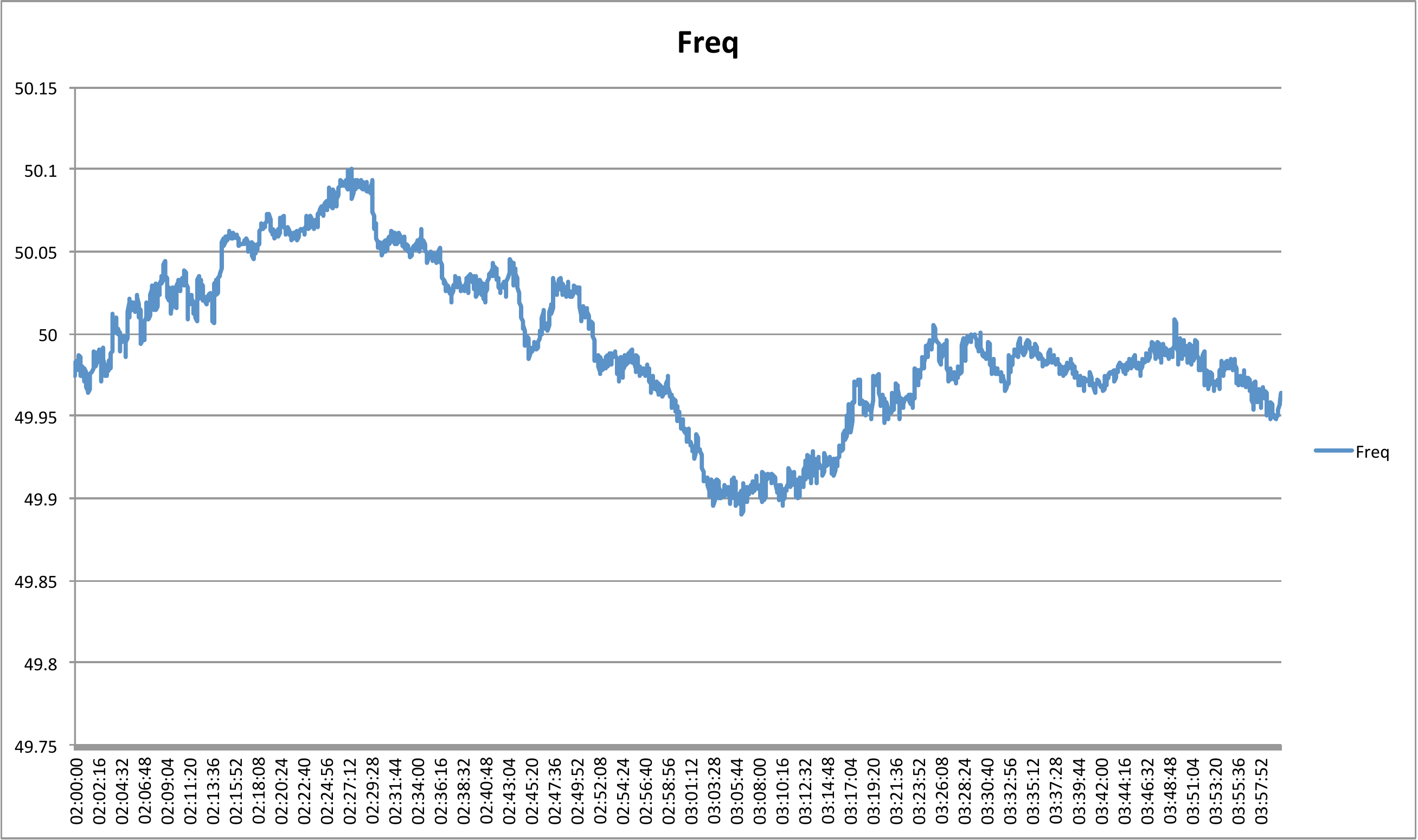} 
   \caption{A frequency trace over 2 hours from National Grid \cite{NG}.}
   \label{fig:freqtrace}
\end{figure}

One could expect the most important mode behaviour to be an OU process for $f_o$, assuming $o$ to be a central node for the network.  Indeed, using GPML, I found that a 2-hour trace of frequency at 1-second intervals, Figure~\ref{fig:freqtrace}, which was publicly available from National Grid \cite{NG}, fit reasonably well to an OU process with a decay time of about 30 minutes and amplitude 0.045Hz.  The time constant is so long compared to the period (about 2 seconds) or decay time (about 20 seconds) of typical inter-area oscillations that it is hardly relevant, and one could just say that the basic behaviour of $f_o$ is a Wiener process (random walk) rather than OU.  The inferred decay time is a significant fraction of the duration of the time series, so might not be determined very accurately.

\begin{figure}[htbp] 
   \centering
   \includegraphics[width=4in]{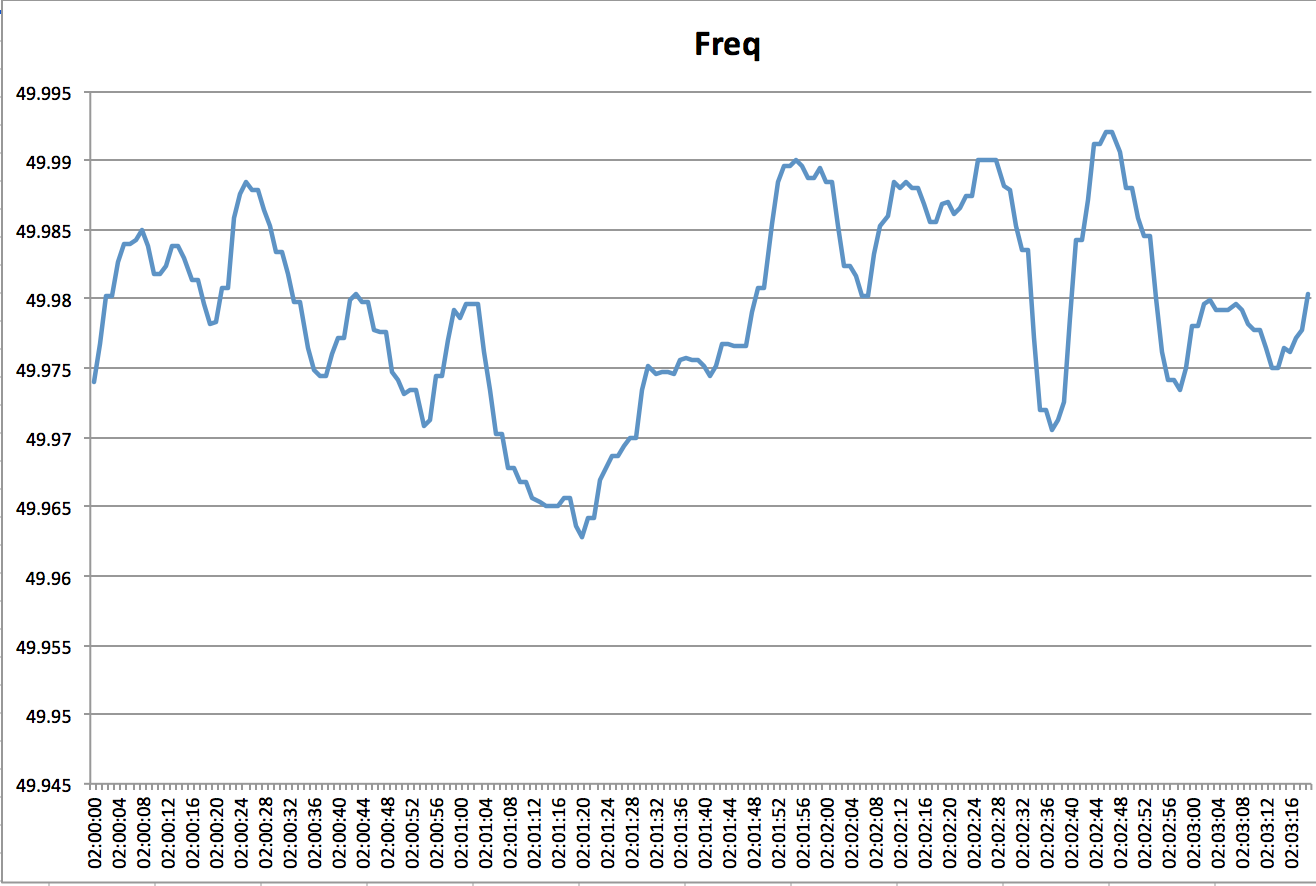} 
   \caption{The first 3 minutes 20 seconds of the frequency trace.}
   \label{fig:short}
\end{figure}

\begin{figure}[htbp] 
   \centering
   \includegraphics[width=4in]{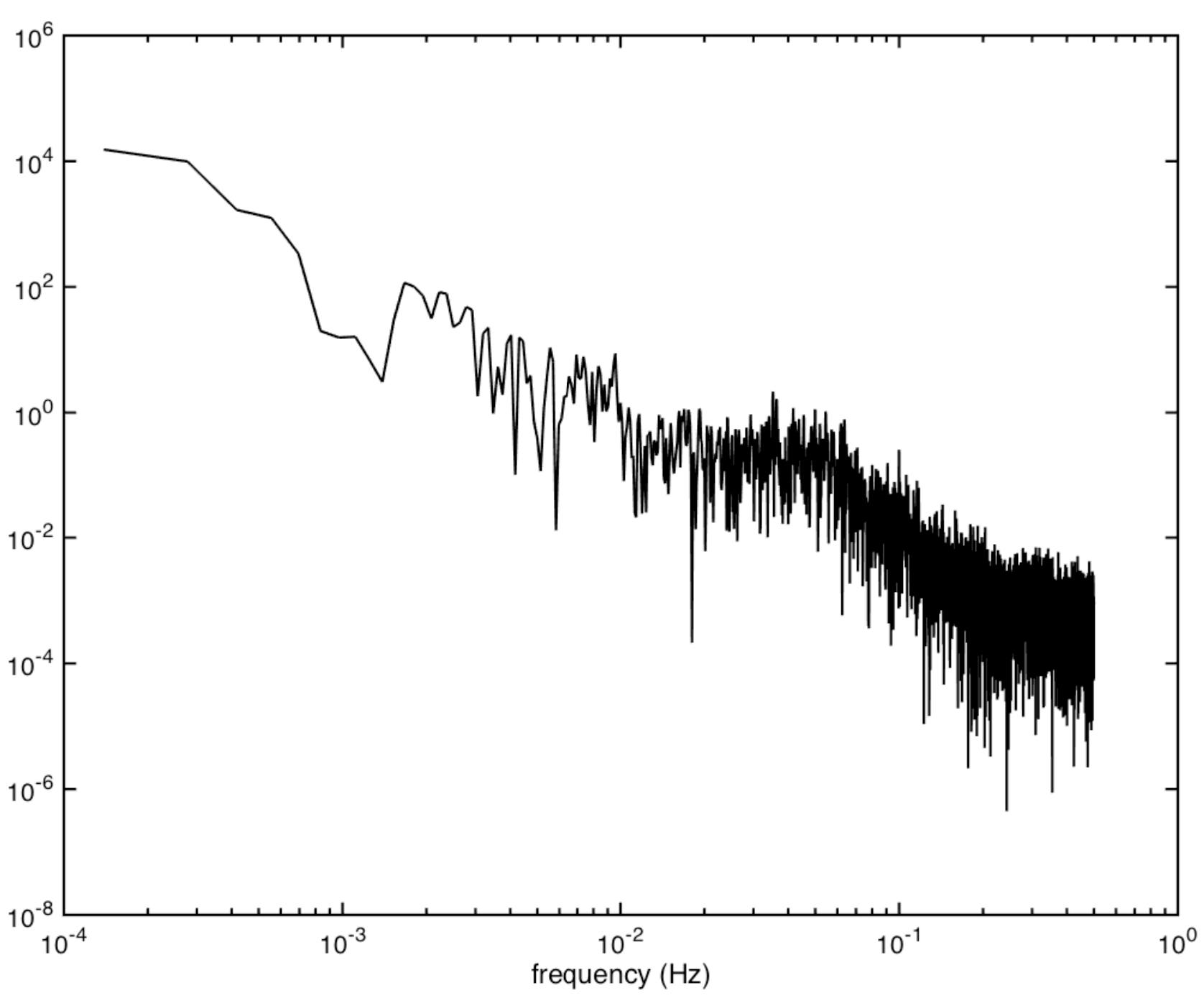} 
   \caption{Loglog plot of the power spectrum of the data of Figure~\ref{fig:freqtrace} using a Hann window.}
   \label{fig:pspec}
\end{figure}

On shorter timescales, however, the data look differentiable (Figure~\ref{fig:short}).  This is my principal reason for rejecting the hypothesis (e.g.~\cite{WBT}) that power imbalance is a white Gaussian noise, because that would make frequency a nowhere differentiable function of time.  Instead I propose that power imbalance is a first-order filtered white Gaussian noise.  Analysis of the power spectrum of fluctuations in the frequency support this proposal.  Figure~\ref{fig:pspec} shows a loglog plot of the power spectrum of the data of Figure~\ref{fig:freqtrace} multiplied by a Hann window function ($\sin^2(\pi t/T)$, where $T=7200$ sec is the duration of the series) to prevent the jump between the values at the two ends provoking high frequency components.  The main part of Figure~\ref{fig:pspec} has a slope near $-2$, consistent with frequency being an OU process.  But for frequency larger than $0.04$Hz (period 25 seconds) the slope steepens, plausibly to $-4$, until the inevitable fact that the data was provided at only 1 second intervals causes a flattening off of the power spectrum at the Nyquist frequency of $0.5$Hz.  National Grid have the data at $1/50$sec intervals, but that is confidential so I can't use it here.  Otherwise we could see if the slope $-4$ extends to higher frequency.

\begin{figure}[htbp] 
   \centering
   \includegraphics[width=3in]{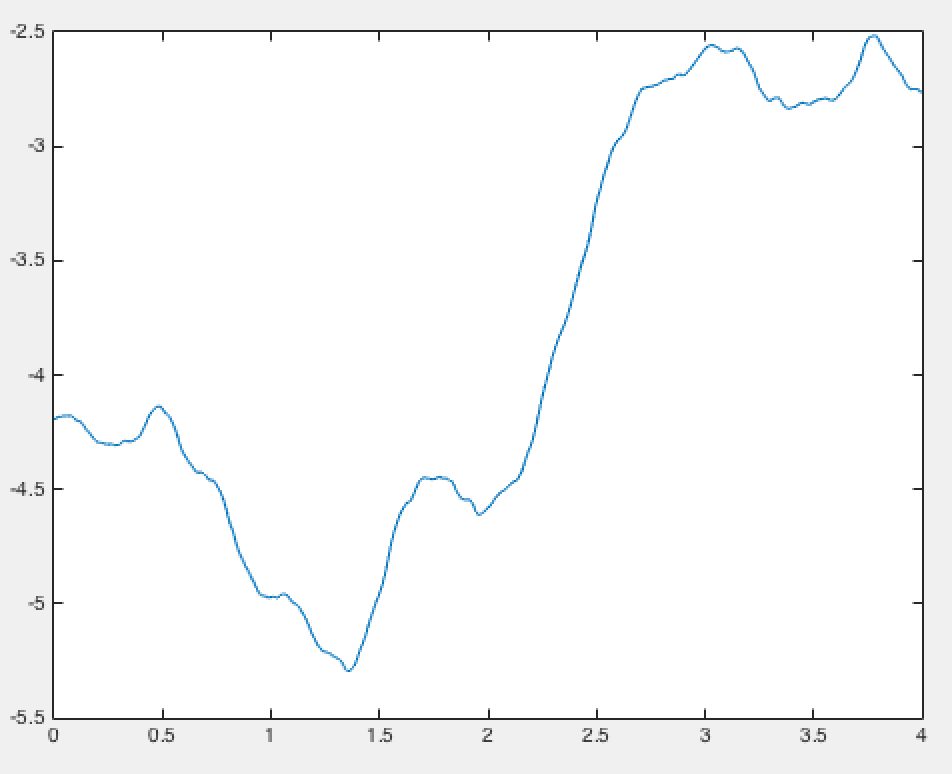} 
   \caption{A sample from the filtered OU process for $\Gamma=1/e$, $J=e^2$}
   \label{fig:fou}
\end{figure}

A simple model for the data is a first-order filtered OU process (FOU).  To justify this, imagine the system is aggregated to a single node.  Then we have two equations of the form
\begin{eqnarray}
M \dot{\delta f} &=& -\gamma \delta f + \delta p \\
\dot{\delta p} &=& -J \delta p + \sigma \xi. \nonumber
\end{eqnarray}
It follows from the second equation that $\delta p$ is OU with covariance function $k(\tau) = \frac{\sigma^2}{2J} e^{-J|\tau|}$.  Then applying (\ref{eq:covt}) we see that $\delta f$ is a GP with covariance function 
\begin{equation}
C(\tau)=\int_0^\infty ds \int_{-\infty}^{\tau+s} d\tau' h(s) k(\tau') h(\tau+s-\tau'),
\end{equation}
where $h$ is the impulse response for the first equation, viz.~$h(s) = \frac{1}{M} e^{-\Gamma s}$, with $\Gamma = \gamma/M$.  Computation of the integral (for the generic case $\Gamma \ne J$) yields
\begin{equation}
C(\tau)=\frac{\sigma^2}{2JM\gamma(\Gamma^2-J^2)}(\Gamma e^{-J|\tau|} - J e^{-\Gamma |\tau|}).
\end{equation}
A sample from the FOU process is shown in Figure~\ref{fig:fou}.
Note that the same covariance function arises for the overdamped linear Langevin process, with $-\Gamma$ and $-J$ being the two real eigenvalues.  

Fitting an FOU to the 2 hours of data with GPML yields time constants $1/\Gamma$ and $1/J$ around $11.1$mins and $1.87$secs, though one can not say from the data analysis which is which (that is an interesting challenge).
It is again awkward that the data is not available at more frequent intervals than 1 second, as the determined time constant 1.87 sec is close to this limit.  A more thorough treatment would evaluate the posterior uncertainty in the parameter fits and attempt to resolve the discrepancies between the previously estimated OU time constant of 30 mins and the current one of 11.1 mins, and between the eyeball estimate $f=0.04$Hz from Fig~\ref{fig:pspec} of where the slope changes, giving a time constant of $(2\pi f)^{-1}$ around 4 sec, and the current one of 1.87 sec.

Over long timescales, deviations from Gaussianity have been established \cite{SBAWT}.  Nevertheless, I believe this does not invalidate Gaussian modelling for short times.

To take this project further, we need next to tackle how a typical two-node system behaves.  This would be the simplest system that could show an inter-area oscillation.  It needs data for the phase difference between the two nodes and their frequencies, and it needs the Kalman filter coding up for at least five dimensions (2 frequencies, 1 phase difference, 2 power imbalances).





\section{Discussion}

I have presented a method to detect oscillations in systems with many components.
It is promising because it can integrate data from many locations simultaneously to enhance the sensitivity of detection of modes of oscillation, and it can run in real-time with constant computation time per observation.

\cite{GDB} consider the problem of calculating modes and mode shapes from phasor measurement units (PMU) in an AC electrical network to have been solved.  They cite \cite{PTD, ZTPM, MV, BPTM}.  I am not so convinced.  I think it would be good to try the method of this paper on that problem.

Detection of modes of oscillation is important in many other contexts.  One example is to detect soft (i.e.~lightly damped) modes for civil engineering structures such as buildings and bridges, e.g.~Ch.13 of \cite{HF}.  Another is the identification of modes of oscillation in the sun (helioseismology), which enables to deduce its temperature and rotation profiles \cite{Ko}.  A third is the analysis of gene expression data, e.g.~\cite{PMPR}.  A fourth is the analysis of business cycles, e.g.~Ch.4 of \cite{Rom}, which have been seen for a long time but are still not understood.

Detection of oscillations is a very old subject, so we next give a brief review of traditional methods.

A standard approach to detecting oscillations is to identify peaks in the Fourier spectrum \cite{HF} or variants \cite{BZA}.  
For example, the response $x$ of the second-order system
\begin{equation}
m\ddot{x}+\beta \dot{x} + kx = \eta
\label{eq:second}
\end{equation}
to noise $\eta$ with power spectrum $P$ has power spectrum 
\begin{equation}
|\hat{x}(\Omega)|^2 = \frac{P(\Omega)}{(k-m\Omega^2)^2 + \beta^2 \Omega^2}
\end{equation}
as a function of frequency $\Omega$.  So if the noise is white ($P$ is constant), then the inverse quality factor $Q^{-1} = \frac{\beta}{\sqrt{mk}}$ is precisely the fullwidth at half maximum for the power spectrum $\Omega^2 |\hat{x}(\Omega)|^2$ of the velocity $\dot{x}$ (its maximum is at $\Omega_{res} = \sqrt{k/m}$, known as the resonant frequency), and the damping ratio $\zeta = \frac12 Q^{-1}$ is the halfwidth at half maximum.  For $P$ slowly varying on the scale of $\frac{\beta}{\sqrt{mk}}$, the results remain good approximations.
This was given a sound grounding in Bayesian analysis (see \cite{Gre} for a survey and \cite{B} for a pedagogical presentation), but still suffers from issues like dealing with trends, choosing windowing functions, missing data, failure to cater for slowly shifting phase, and poor theoretical justification for taking more than the largest peak if one wants to infer more than one mode of oscillation.  
 
Wavelet transforms are popular for resolving signals in both time and frequency (up to the limits of the uncertainty principle), but I am not aware whether they can give an estimate of damping rate.

Another approach is to study the effect of an impulse (the Prony method and variants like MUSIC and ESPRIT, e.g.~\cite{PLH}), but many real-world systems may not be subjectable to impulses.  For a review of these and some other methods (e.g.~Hilbert transform), see \cite{ZD}.\footnote{As yet another method, I learnt back in the mid-1980s that a good way to determine the eigenvalues of an asymptotically stable system from the response to an impulse is to Laplace transform the response numerically and then fit a Pad\'e approximation and read off its poles.}
One defect of the approach is that the forcing might not be Gaussian.  For example, even a Poisson process with independent Gaussian amplitude is not Gaussian.  Indeed, a consequence of the Gaussian assumption is that the covariance of the response is time-symmetric, whereas this may not be true for real systems.  As already mentioned, evidence for Gaussian distribution of electrical load is given in Fig.14 of \cite{TT+}, but they do not report on time-correlation.  Load variations are likely to be independent, however, which would make them Gaussian and white.  On the other hand, wind power is unlikely to be delta-correlated.  There is considerable research on the statistics of wind power, e.g.~\cite{DPP, TWD, WBCF}.

Another defect of the approach is that it does not allow for nonlinearity.  Nevertheless, for small fluctuations around an equilibrium, linearising is a good approach.  It will fail to give a good approximation, however, if the eigenvalues of any mode approach or cross the imaginary axis.   A big question with power flow oscillations, gene expression and business cycles is whether there is a limit cycle of some underlying deterministic dynamics, or just lightly damped oscillations around an equilibrium forced by noise.  Figure~\ref{fig:TR+} suggests to me that there was a Hopf bifurcation, but the common wisdom in the power system community is that it was just a large kick that set off a lightly damped mode of oscillation.
For gene expression this has been addressed by \cite{D+}.  For business cycles, most economists decided long ago that they are just a near unit root process (meaning lightly damped oscillations forced by shocks) \cite{Rom}, though Grandmont proposed deterministic models with a variety of forms of dynamics \cite{Gra}.  \cite{Sim} fitted a VAR model, but with perhaps too many free parameters.  Our approach would restrict to a small number of modes.

An interesting issue is that if the noise is considered to be the result of filtering white noise then our method also finds the modes of the filter.  Without further information about the structure of the system or direct observations of the forcing process, we see no way of distinguishing between modes of the filter and modes of the system from observations of just the system.  An example of this was given in Section~\ref{sec:ACelec}.  

To detect periodic components, my brother David \cite{M} proposed the family of stationary covariance functions of the form
\begin{equation}
k(t) = \sigma^2 \exp \left(-\frac{2 \sin^2(\omega t/2)}{\lambda^2} \right), 
\end{equation}
for which samples are exactly periodic with period $2\pi/\omega$.
A slight modification was used in \cite{L+} to remove the effect of its non-zero mean, namely
\begin{equation}
k(t) = \sigma^2 \frac{\exp(\lambda^{-2} \cos \omega t) - I_0(\lambda^{-2})}{\exp(\lambda^{-2}) - I_0(\lambda^{-2})},
\end{equation}
where $I_0$ is a modified Bessel function of the first kind.
It has the limiting form
\begin{equation}
k(t) = \sigma^2 \cos(\omega t)
\end{equation}
as $\lambda \to \infty$, called the Cos kernel, which has the property that it forces anti-periodicity with anti-period $\pi/\omega$: $f(t+\pi/\omega) = -f(t)$.  Although these have found valuable uses, and can be made less rigid by multiplication by a decaying kernel such as $\exp(-\alpha |t|)$ (which with the Cos kernel produces OUosc), it seems to me highly preferable to start from the point of view of a linear system forced by noise.

I conclude with a suggested improved approach to nuclear magnetic resonance imaging.  The present standard approach is to apply an electromagnetic pulse that simultaneously excites all the single-quantum NMR transitions.  The resulting time-domain signal is Fourier transformed to reveal NMR absorption intensity against frequency.  I suggest instead to apply electromagnetic noise and from the resulting response to infer the frequencies and damping rates by the Gaussian processes of this paper.

\section*{Acknowledgements}

I am grateful to Ben Marshall of National Grid for proposing the problem of detecting inter-area oscillations in Jan 2015, and to him and his colleague Phillip Ashton for helpful discussions on the topic and pointers to the literature; to MSc student Tajhame Francis for initial investigations by spectral analysis; to my brother David for telling me to ``Use a Gaussian process"; to PhD student Marcos Tello Fraile and postdoc Lisa Flatley for trying to follow my suggestions; to Hannes Nickisch and Colm Connaughton for helping me implement my resulting solutions in GPML; to Carl Rasmussen and Hannes Nickisch for having created GPML; to Zoubin Ghahramani for answering some questions about GPs; to Igor Mezic and Yoshihiko Susuki for discussions on modelling AC networks; to undergraduate summer project student John Prater for coding up the $2\times 2$ underdamped linear Langevin covariance for GPML; and to Chris Williams,  Darren Wilkinson and especially Janusz Bialek for useful comments and questions.  The beginning of the work was supported by National Grid under Network Innovation Allowance award NIA\_NGET0161.  The later parts were supported by the Alan Turing Institute under award TU/B/000101.

\section*{References}

\end{document}